\documentclass[runningheads]{llncs}
\usepackage{amsmath}
\usepackage{multirow}
\usepackage{longtable}
\usepackage{tabularx}
\usepackage{graphicx}
\usepackage[caption=false]{subfig}
\usepackage{xcolor}
\usepackage{float}
\usepackage[export]{adjustbox}
\usepackage{wrapfig}
\usepackage{algpseudocode, algorithm}

\usepackage{graphicx}
\newcommand{\vecX}{\mathbf{x}}
\newcommand{\vecP}{\mathbf{p}}
\newcommand{\vecG}{\mathbf{g}}
\newcommand{\vecE}{\mathbf{e}}

\begin{document}
\title{Unsupervised anomaly detection for discrete sequence healthcare data}
%
%
\author{Victoria Snorovikhina\inst{1} \and
Alexey Zaytsev\inst{1}}

%

%
\institute{Skolkovo Institute of Science and Technology, Moscow, Russia 121205 \and
\institute{Skolkovo Institute of Science and Technology, Moscow, Russia 121205}}
\maketitle              
\begin{abstract}
Fraud in healthcare is widespread, as doctors could prescribe unnecessary treatments to increase bills. Insurance companies want to detect these anomalous fraudulent bills and reduce their losses. Traditional fraud detection methods use expert rules and manual data processing. 

Recently, machine learning techniques automate this process, but hand-labeled data is extremely costly and usually out of date. We propose a machine learning model that automates fraud detection in an unsupervised way. Two deep learning approaches include LSTM neural network for prediction next patient visit and a seq2seq model. For normalization of produced anomaly scores, we propose Empirical Distribution Function (EDF) approach. So, the algorithm works with high class imbalance problems. 

We use real data on sequences of patients' visits data from Allianz company for the validation. The models provide state-of-the-art results for unsupervised anomaly detection for fraud detection in healthcare. Our EDF approach further improves the quality of LSTM model. 

\end{abstract}
\section{Introduction}
Healthcare is an essential part of modern society, and the modern medical system is one of the main achievements of humankind. 
However, both private healthcare companies and government healthcare systems face fraudulent cases every day, and this number keeps increasing every year. 
Clinics as service providers prescribe unnecessary expensive medications and 
procedures. 
Moreover, a patient and a doctor can falsify a patient`s diagnosis to get money for medical services.
Insurance companies have to cover such excessive bills and want to detect these fraudulent expenses. 

Traditionally detection of such frauds was a manual routine for expensive subject area experts~\cite{article_2}, but now since machine learning techniques and deep learning tools become a natural part of business processes, automatic fraud detection systems were built~\cite{Bauder2017}. 

\begin{figure}[h]
\begin{minipage}[h]{1\linewidth}
\centering
\includegraphics[width=0.75\linewidth]{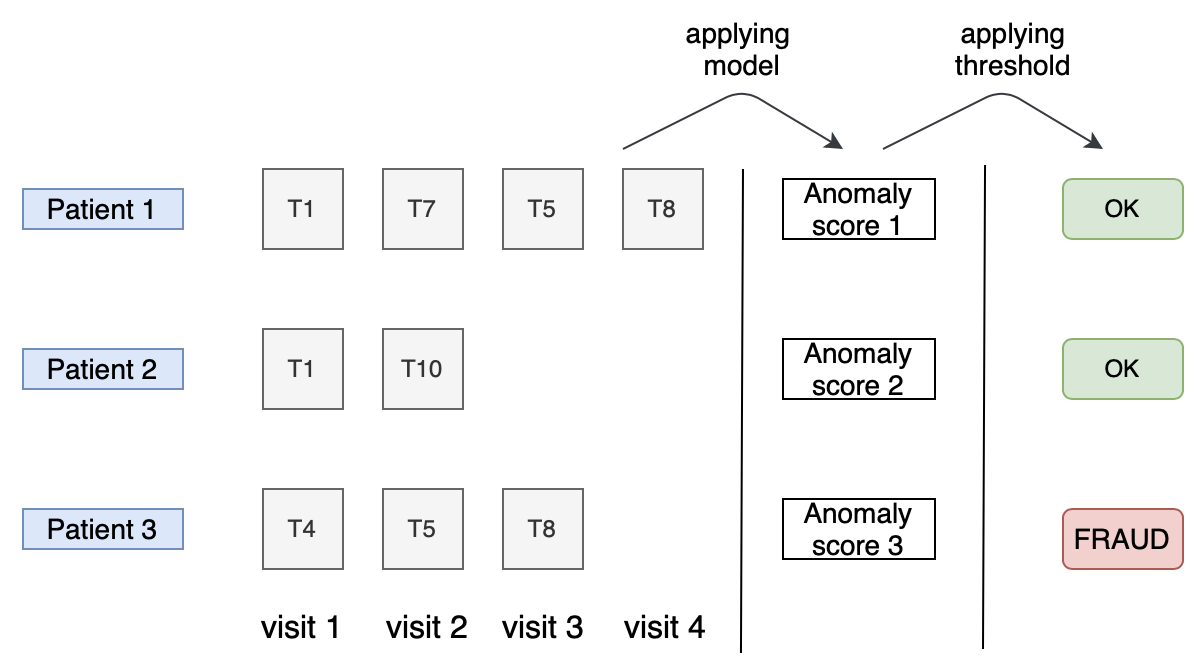}
 \caption{Pipeline of a proposed solution. For each patient, we have information on a code of prescribed treatment for each visit. We recover these treatments with a generative model and get anomaly scores on the base of errors of our generative model. Then if the anomaly score if higher than a selected threshold, we signal about a fraud. Our generative model can deal with sequences of various length and treatments from a dictionary of a large size.}
 \label{Figure:scheme}
  \end{minipage}
\end{figure}

A machine learning guided fraud detection is faster and requires lower human involvement, but the solution is not ideal, as the problem itself is hard. 
The typical approach is to hire experts to obtain labeled data~\cite{article_1} and then construct a classification model from an available imbalanced dataset with many fair and a few fraudulent records. 
Due to a large amount of data and complicated fraud patterns, only experienced auditors are able to detect fraudulent cases; thus, data collection is expensive. Also, machine learning models are able to catch only identified types of frauds.
Moreover, as the resulting dataset is imbalanced, we have to carefully construct machine learning involving methods aimed at the solution of imbalanced classification problem~\cite{branco2016survey,kozlovskaia2017deep}.

Unsupervised fraud detection systems can successfully deal with these issues.
For example, the authors in~\cite{article_1} identify if a particular doctor conducts fraud or not using an open dataset. They assume that doctors with common specialties behave in the same way with similar average bills and medicine price rate~\cite{Bauder2017}. So, these types of models help to detect the doctor, who is prone to fraud. 
The work~\cite{article_2} detect frauds at the patient level using the private dataset, as there are no open data for this problem. 
The model takes general information about a patient as input: the number of medical procedures that were provided, the average procedure bill, and so on.
So, the existing approaches in healthcare exist but utilize only hand-crafted features \cite{article_2}, thus not being able to detect frauds on the base of complex semi-structured data. 

In other areas, anomaly and fraud detection methods are wider, and can roughly be divided into four directions. The first direction considers both classic and neural network supervised~\cite{Bauder2017} and unsupervised Machine Learning algorithms \cite{article_9}. The second direction considers various probabilistic approaches~\cite{p41} and relies on an approximation of the generative distribution of the observed data. The third direction adopts autoencoder models \cite{wiewel2019continual} and learn data representations using sequence to sequence (seq2seq) architectures. The hybrid models also exist~\cite{zong2018deep}. Recently, the state of the art approaches for anomaly detection based on sequence data are autoencoder models~\cite{zimmerer2018context} and recurrent neural networks (RNN) \cite{article_8}.

This work advances unsupervised anomaly detection for healthcare semi-structured data. 
We deal with discrete sequence variables and modify existing anomaly detection approach to handle complex data from large dictionaries. 
The main idea is to reconstruct a sequence of treatment using a generative model and compare it to the initial one, treating reconstruction error values as anomaly scores.
To signal about an anomaly, we apply a threshold to these scores. 

The raw data consist of semi-structured sequences of treatments for different patients, plus additional features like patient`s age, sex, etc. Treatments are coded, so it could be interpreted as a set of pre-defined tokens.
The pipeline of the proposed solution could be seen in Figure~\ref{Figure:scheme}.

To sum up, our contributions are the following:
\begin{itemize}
    \item We apply unsupervised anomaly detection to fraud detection based on healthcare records.
    \item We adopt a classic anomaly detection approach for regression to a classification problem with a generative model for sequences of treatments.
    We consider the local LSTM model for the prediction of a single token and a sequence to sequence model based on LSTM to recover the whole sequence.
    \item For the first model, we provide a new normalization procedure to handle a large dictionary size about 2000 and thus an imbalanced classification problem with a large number of classes.
\end{itemize}

The paper is organised as follows: In Section~\ref{sec:related_works}, related work on anomaly detection in sequences (especially in a healthcare) is summarised. Section~\ref{sec:methods} describes the proposed solution, models and approaches of errors definition. Approach to handle sequence imbalance is also in this section. 
Section~\ref{sec:experiments} is devoted to machine experiments with real data. Finally, Section~\ref{sec:conclusion} concludes the paper.

\section{Related works}
\label{sec:related_works}

There are two types of works related to the problem at hand:
fraud detection in healthcare and anomaly detection in general, especially anomaly detection for semi-structured sequences. 
For a general review of anomaly detection in industry and healthcare look at~\cite{habeeb2019real}, for a recent survey on applications of deep learning to unsupervised anomaly detection, \cite{kwon2019survey,chalapathy2019deep} and \cite{kiran2018overview} can be useful.
Here we present the part of the research that we believe is the most relevant to our studies.

In healthcare, there is a widely-used open dataset \emph{Medicare claims data}~\cite{christopher2003national}, which includes aggregated information by doctors and patients. The data have labeling of doctors: do they fraud or not?
In \cite{article_1} authors used Logistic Regression and Random Forest to work with this dataset. 

In~\cite{article_2}, the authors focus on supervised detection of upcoding fraud, when doctors replace code for an actual service with a code of a more expensive one. For example, a procedure that lasted fifteen minutes can be coded as a more expensive thirty minutes visit. Used data consisted of a sequence of coded visits. 

The paper~\cite{Bauder2017} considers unsupervised approaches to healthcare fraud detection.
In particular, the authors investigate the applicability of k-nearest Neighbors, Mahalanobis distance, an autoencoder, and a hybrid approach based on a pre-trained autoencoder without labeled data as input for supervised classifiers. 
In \cite{article_13}, authors use Generative Adversarial Network to detect anomalies for healthcare providers.

Applications of supervised deep learning models also attract attention in deep learning models~\cite{farbmacher2019explainable,article_14}. 
The authors in~\cite{article_14} used embedding techniques and both classical Gradient Boosting and deep learning approaches. 

For unsupervised anomaly detection in general, there are clustering techniques~\cite{article_9}: authors used the Isolation Forest algorithm, which is considered as one of the most popular and easy in the usage of anomaly detection algorithms. See also usage of probabilistic approaches in~\cite{p41}, usage of sequence to sequence architectures in \cite{wiewel2019continual}, and usage of hybrid models in~\cite{zong2018deep}. 

In~\cite{fca}, authors have investigated a problem of human trafficking, which requires detection and interpretability, so they applied used Formal Concept Analysis. In~\cite{fca_2}, authors also have investigated this algorithm, but for specific sequences. In ~\cite{fca_mark}, authors used Hidden Markov models in a healthcare field, which provided better results than FCA.

In \cite{article_8}, the authors used the LSTM to predict each subsequent measurement of the spacecraft. They proposed an automatic threshold selection to determine an anomaly, indeed due to the mean and standard deviation of LSTM errors. It is worth to mention that this domain of study is using raw data as input to a neural network, which provides better accuracy compared to a model based on processed data.

We see that no one proceeds semi-structured medical insurance data to detect fraud in an unsupervised manner.
Moreover, general machine learning literature lacks methods that can deal with a moderate token dictionary size for the anomaly detection problem and, in particular, automatically select multiple thresholds for a base anomaly score for each considered class label for a dictionary.
\section{Methods}
\label{sec:methods}

\subsection{General scheme}

We have a set of size $n$ of patients. 
Every patient $i$ is represented as a sequence of observations $X_i = \{\vecX_{1i}, \vecX_{2i}, \ldots, \vecX_{T_i i}\}$, $t \in \overline{1,T_i}$.
The total number of visits for a patient is $T_i$.
Each vector $\vecX_{ji}$ is the description of a particular $j$-th visit of $i$-th patient.
$\vecX_{ji}$ consists of treatment type from a dictionary of size $d_t$, cost type from a dictionary of size $d_c$ and benefit type from a dictionary of size $d_b$.
We also pass the general information $\mathbf{g}_i$ about the patient at each sequential step.

\begin{figure}[t]
\centering
\includegraphics[width=0.6\linewidth]{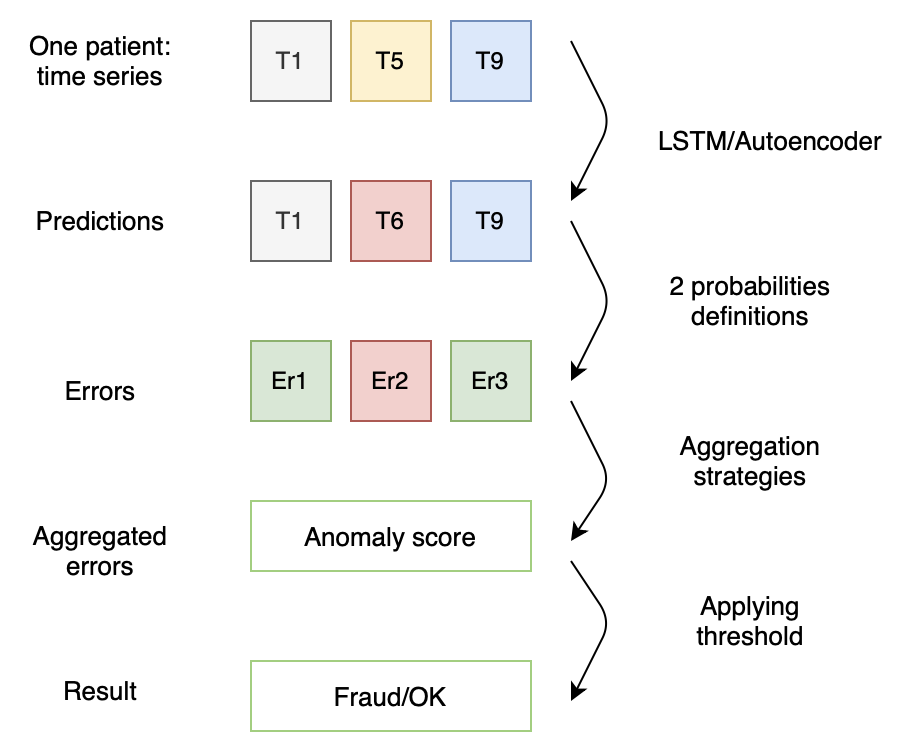}
 \caption{General scheme of the proposed solution. We predict a sequence using a generation model and obtain errors for each token. Then we get a general anomaly score on the base of aggregation of these errors. Applying a selected threshold, we can say if a particular sequence of tokens is a fraud or not and identify fraudulent tokens as tokens with highest errors.}
 \label{Figure:pipeline}
\end{figure}

The pipeline of a proposed solution is in Figure \ref{Figure:pipeline}.
Below we provide more details of each step from this pipeline.

In order to detect whether a particular sequence has fraud visits or not, we will measure the likelihood of the sequence $p(X_i)$ using a seq2seq approach that we call the Autoencoder model and a token by a token approach which we call LSTM model.
To do this, we either pass $X_i$ through the seq2seq model and get probabilities for each token in output $\mathbf{p}_{ij}$ or predict a token using all previous tokes for LSTM model to get another vector $\mathbf{p}_{ij}$.
Then we calculate the likelihood of a particular token $j$ using the following formula: $e_{x_{ij}} = 1 - p_{x_{ij}}$, if $x_{ij}$ is a true label token or $e_{x_{ij}} = p_{x_{ij}}$, if $x_{ij}$ is a false label token, where $e_{x_{ij}}$ is an error. 
We recover through an autoencoder only a part of sequence related to treatments or treatment types.
Thus, we provide results of experiments for large (treatments) and small (treatment types) number of classes (tokens).


To get an estimate of a sequence likelihood
we, first, built either vector of errors or a matrix of errors. Vector corresponds to the case, where errors on only true token labels are taking into account, the matrix is consist of errors both on true and false token labels. Secondly, we use sum/max pooling to get a single anomaly score for a sequence.
The final prediction is , we compare the obtained score with a threshold.
We select a threshold to get the recall $0.8$. 
This is the only number we calibrate using fraud labels, we state that our approach is unsupervised.

\subsection{Sequence models}

\paragraph{\textbf{LSTM model}}~\cite{lstm} is the most widely used type of Recurrent neural networks (RNNs) \cite{Sherstinsky_2020}. 
We use LSTM architecture to get the probabilities for the next treatment $\vecP_{ij}$ on the base of previous information $ \{\vecX_{1i}, \ldots, \vecX_{(j -1)i} \}$. 
The architecture works as follows for each step $j$:
\begin{enumerate}
    \item Embed treatment type, cost type, and benefit type using separate embedding layers with trainable embedding matrices $E$ of size $d \times e$, where $d$ is the dictionary size, and $e$ is the embedding size. Embedding size is one for all feature types. For treatments embeddings size is 128, for treatment type it equals 32. Concatenate these embeddings and the general information about a customer~$\vecG_i$.
    \item Pass this concatenated vector to two successive LSTM blocks.
    \item Pass the resulting hidden state to a linear layer to get probabilities of each token $\vecP_{ij}$
\end{enumerate}

By applying this model for each token of initial sequence we get a set of vector of probabilities $P_i = \{ \vecP_{ij} \}_{j = 1}^{T_i}$

\paragraph{\textbf{Autoencoder model}} is a sequence-to-sequence architecture~\cite{il}. 
We learn the model to copy a sequence, such that the generated sequence is as close to initial as possible.
The intuition is that the network learns the representation of sequences structure, so it would be difficult to recover fraudulent sequences with unexpected treatments inside. 

The model consists of an encoder and a decoder. The encoder constructs a representation of an input sequence $\mathbf{r}_i = E(X_i)$ that equals to the hidden state of the last recurrent block. 
The decoder tries to generate the initial sequence for the representation: $X'_i = D(\mathbf{r}_i)  \approx X_i$.
As a result, the model outputs the probability distribution for every element of a sequence.   

We used a bi-directional LSTM network \cite{10.1109/78.650093} as encoder, unidirectional LSTM network as a decoder. Both had two layers and embedding sizes 128. We also used a context attention vector~\cite{Lai2019HumanVM}. Every decoder hidden state is passed through a dense layer by applying a soft-max function, and we obtain probability distribution for the next treatment.

\begin{table}[ht]
    \centering
    \begin{tabular}{ll}
        \hline
         \textbf{Feature} & \textbf{Description}\\
        \hline
         Treatment& 2204 unique values\\
         Treatment type & 17 unique values (aggregated treatments)\\
         Treatment number& Prescribed number of each treatment\\
         Factor & Factor of the treatments` amount\\
         Cost & Cost of a particular visit\\
         Cost Type & 11 cost categories\\
         Benefit Type& 24 treatments` combinations types\\
         \hline
    \end{tabular}
    \caption{Features for the description of each visit of a patient}
    \label{table:data}
\end{table}

\subsection{Anomaly score}

Given this probability distribution $p(X_i)$ either from LSTM or Autoencoder models, let us define anomaly score.

For every patient the output of a model is a set of vectors of probabilities $P_i = \{\vecP_{ij}\}_{j = 1}^{T_i}$. Length of vectors $\vecP_{ij}$ equals to the size of the dictionary of treatment $d$. 

Given true labels, we define error as one minus the probability of a true label: $e_{ij x_{ij}} = 1 - p_{ijx_{ij}}$, where $p_{ijx_{ij}}$ is $x_{ij}$-th element of probability vector that corresponds to the index of the true token label $x_{ij}$.
We calculate errors that correspond to high probabilities of false labels as $e_{ijk} = p_{ijk}$, $k \ne x_{ij}$.

If we concatenate all errors for true classes $e_{ij x_{ij}}$ we get a vector of errors $\vecE_{i} = \{ e_{ij x_{ij}} \}_{j = 1}^{T_i}$.
If we concatenate all errors for all classes we have a matrix of errors $E_i = \{e_{ijk} \}_{j = \overline{1, T_i}; k=\overline{1, d}}$, where $d$ is the dictionary size.

To get a single anomaly score $a_i$ from a vector or a matrix, we aggregate them using pooling.
We consider sum pooling and max pooling, 
as in our experiments, mean pooling worked worse.
For the vector aggregation we get:
\[
a^{\mathrm{sum}}_i = \sum_{j = 1}^{T_i} e_{ij x_{ij}}, \,\,    
a^{\mathrm{max}}_i = \max_{j = \overline{1, T_i}} e_{ij x_{ij}}.
\]
For the matrix aggregation, we use $E_i$ instead of $\mathbf{e}$ to extract sum or maximum.

\begin{figure}[ht]
\centering
\includegraphics[width=0.8\linewidth]{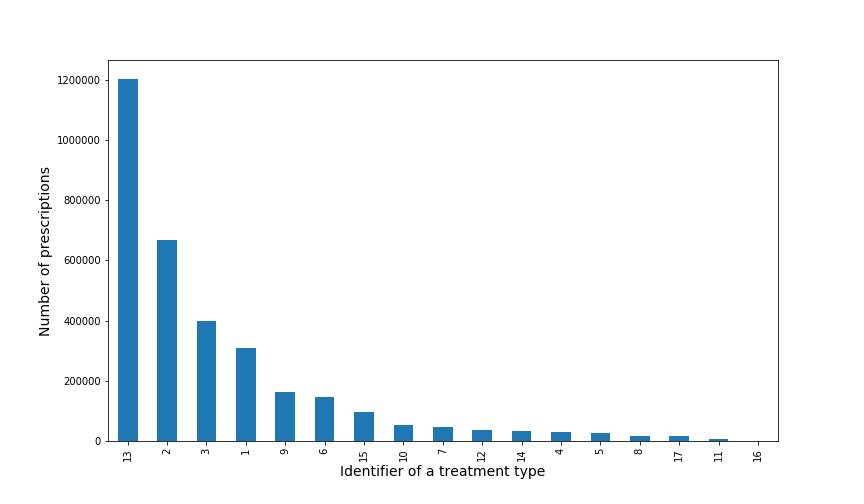}
\caption{Distribution of treatment types}
\label{Figure:tr_types_distr}
\end{figure}

\subsection{EDF approach}

To normalize the probability scores and get meaningful aggregation, we transform the error scores based on their empirical distribution function (EDF). 
EDF is an approximation of a theoretical distribution function, based on an observed sample for a random variable. Assume, there is a sample of $n$ independent real values with common distribution function $\mathbf{e} = \{e_i\}_{i = 1}^n$, then EDF value for a particular $e$ is number of elements in the sample that is smaller than $e$ divided by the sample size $n$:
\[
    \mathrm{EDF}(e) = \frac{1}{n} \sum_{i = 1}^n [e_i < e],
\]
where $[\cdot]$ is the indicator function.

Thus, as cumulative distribution function defines the probability of a variable, EDF defines the relative frequency of a particular point. Therefore, having calculated EDF for errors of every class separately, it provides a better understanding of which points are anomalous. 

We construct $d$ Empirical Distribution Functions for each element using a separate validation sample not used during training. Then we transform errors by replacing error values with EDF values for the corresponding label and get a normalized vector $\hat{\vecE}_{i}$ or a matrix  $\hat{E}_{i}$. We aggregate
these errors in the next step in a similar way, replacing errors with EDF-normalized errors.

\begin{figure}[t]
\centering
\subfloat[ROC curves of unsupervised fraud detection models with respect to large number of classes (treatments) and small number of classes (treatment types). Best ROC AUC is $0.771$. \label{Figure:roc}]{%
  \includegraphics[width=0.45\textwidth]{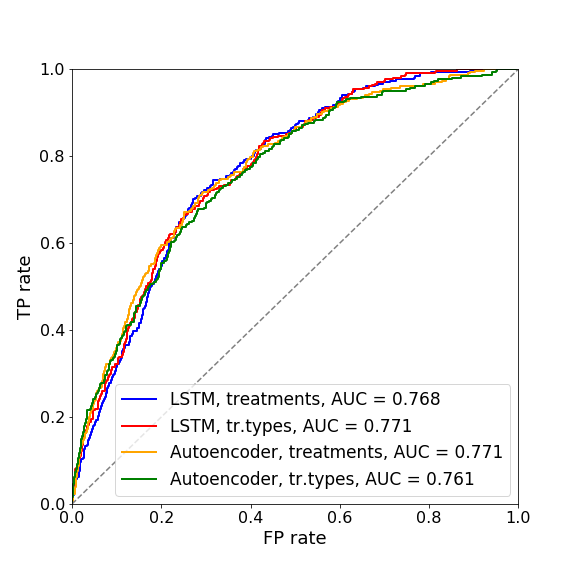}%
}\hfil
\subfloat[PR curves of unsupervised fraud detection models with respect to large number of classes (treatments) and small number of classes (treatment types). Best PR AUC is 0.0720.\label{Figure:pr}]{%
  \includegraphics[width=0.45\textwidth]{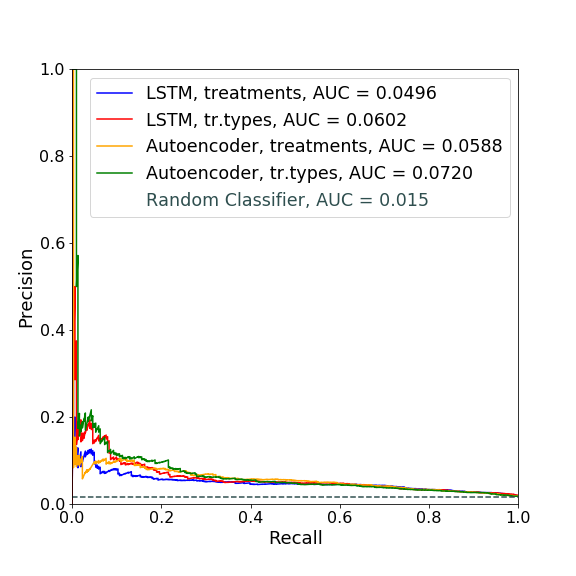}%
}
\caption{Comparison of performance curve for presented algorithms}
\label{fig:performance_curves}
\end{figure}

\begin{table}[th]
    \centering
    \begin{tabular}{lllllll}
        \hline
         Model & \multicolumn{3}{c}{Treatments} & \multicolumn{3}{c}{Treatment types} \\
         & ROC AUC & PR AUC & Precision & ROC AUC & PR AUC & Precision \\
         \hline
         LSTM  &0.743 & 0.0425 & 0.0164 & 0.761 & 0.0681 & 0.0170 \\
         LSTM + EDF  & 0.768 & 0.0499 & {\bf 0.0333} & {\bf 0.771} & 0.0601 & {\bf 0.0325} \\
         Autoencoder  & {\bf 0.771} & {\bf 0.0588} & 0.0331 & 0.761 & {\bf 0.0720} & 0.0319 \\
         Autoencoder + EDF &0.750 & 0.0483 & 0.0317 & 0.760 & 0.0654 &  0.0319 \\
         \hline
    \end{tabular}
    \caption{Quality comparison of unsupervised fraud detection for LSTM and Autoencoder models. Precision is given for the corresponding recall $0.8$. For LSTM additional normalization provided by EDF is useful, while Autoencoder can capture all information without EDF}
    \label{2_models}
\end{table}

\section{Experiments}
\label{sec:experiments}

We compare our anomaly detectors based on recovery of treatments and treatment types to each other and a baseline for a considered applied problem from healthcare insurance.

\subsection{Data}

The data for the current research was provided by a major insurance company~\cite{article_14}. The dataset consists of $350$ thousand records with anonymous patient`s IDs and target labels (fraud or not) for patients. 
About $1.5\%$ records are fraudulent. 

For each patient, we have general features age, sex, insurance type, and total invoice amount and visit-specific features given in Table~\ref{table:data}.
In our model, visits are coded either as treatments or treatment types. 
In Figure~\ref{Figure:tr_types_distr}, we provide a distribution of treatment types concerning its prescribed frequently: 
the histogram demonstrates a strong class imbalance. 


\subsection{Results}

\paragraph{Metrics.} The problem at hand is an imbalanced binary classification, so we use traditional metrics like ROC AUC and area under precision-recall curve PR AUC, where positive samples are the fraudulent ones. We also use ROC and PR curves, as well as precision and recall. 

\paragraph{Training process.} We conduct experiments with sequences of treatments and treatment types independently. 

Patients have a different number of visits; thus, all sequences were padded with zeros to the closest power of two to an initial sequence length. For the padded elements, a network returns zeros.

The training sample includes $95\%$ of the data, and the test sample includes the remaining $5\%$ of the data. We use $5\%$ of the training sample as a validation set to compare model performance and calculate the EDF function. The test consists of $17000$ for patients with $300$ fraudulent cases. Distribution of classes in validation, training, and test datasets are the same.
   
The training process consists of $100$ epochs for the LSTM model and $70$ epochs for the Autoencoder model. 
We use the Adam optimization algorithm and a cross-entropy loss. An initial learning rate is $0.001$ for LSTM model; $3 \times 10^{-6}$ and $10^{-6}$ for treatments and treatment types respectively for autoencoder model. Exponential learning rate decay with a coefficient $0.95$ is used.
We train the LSTM network in the end-to-end fashion. Parameters at the first iteration are initialized randomly.

The used best hyperparameters come from cross-validation for training data and are given below.
Embedding size is one for all feature types, for treatments, it is 128, and for treatment types, it is 32. Batch sizes are 128 for Autoencoder and 256 for the LSTM model. 
For treatment models, we use sum aggregation for a matrix of errors with EDF for the LSTM model and sum aggregations for a matrix of errors for the Autoencoder model. 
For treatment types, we use sum aggregation for the matrix of errors with EDF for the LSTM model and sum aggregations for a matrix of errors for the Autoencoder model. 

\begin{figure}[tb]
\centering
\subfloat[ROC AUC values with respect to visit sequence length. \label{Figure:roc_len}]{%
  \includegraphics[width=0.4\textwidth]{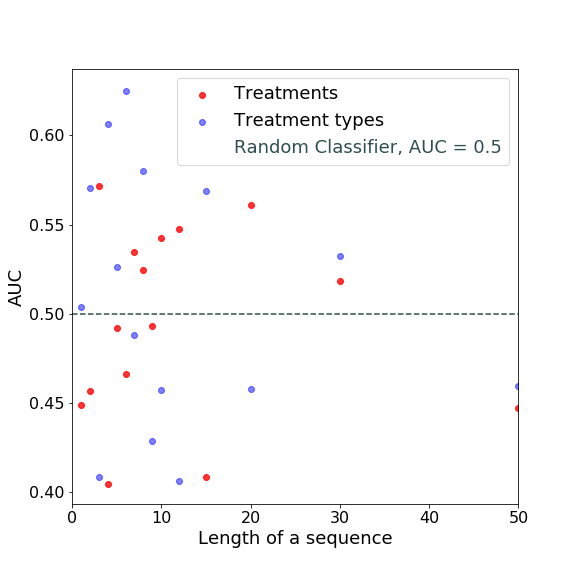}%
}\hfil
\subfloat[PR AUC values with respect to visit sequence length. \label{Figure:pr_len}]{%
  \includegraphics[width=0.4\textwidth]{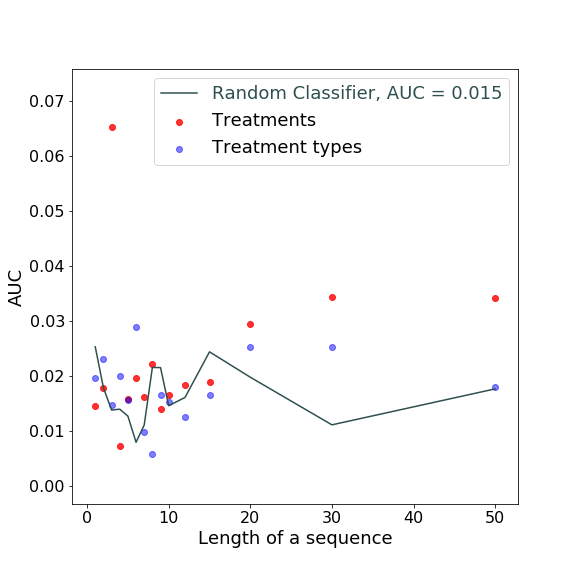}%
}
\caption{Performance of models based on treatments and treatment types for different lengths of sequences. The models work better than a random classifier most of the time.}
\label{fig:length_curves}
\end{figure}

\begin{table}[t]
    \centering
    \begin{tabular}{ccccccc}
        \hline
         & \multicolumn{2}{c}{LSTM} & \multicolumn{2}{c}{Autoencoder}  & 
         \multicolumn{2}{c}{Isolation Forest}  \\
         &&&      &             & \multicolumn{2}{c}{(Baseline)} \\
        & Recall & Precision & Recall & Precision & Recall & Precision \\
        \hline
         Treatments & \textbf{0.80} & \textbf{0.0333} & 0.80 & 0.0331 & 0.07 & 0.08 \\
         Treatment types & 0.80 & 0.0325 & 0.80 & 0.0319 & 0.06 & 0.06 \\
         \hline
    \end{tabular}
    \caption{Comparison of the proposed models and a baseline. We present result for our models LSTM and Autoencoder and for a baseline Isolation Forest. For Isolation Forest we can't reach Recall $0.8$, so we present precision for the maximum possible value of recall. The best combination of precision and recall is marked in bold. Our models are better, than a baseline}
    \label{bas}
\end{table}

\paragraph{Results.} A comparison of quality for both models with and without the EDF approach for the best parameters and aggregation strategies are in Table \ref{2_models}. Precision is calculated with the expected recall 80\%. 
Corresponding ROC and PR curves are in Figures~\ref{fig:performance_curves}, PR curve. 
LSTM and Autoencoder models provide similar quality of anomaly detection with ROC AUC $0.77$. For a large number of classes in LSTM model, the difference in precision are $\sim 1.7$\%; for a small number of classes is $\sim 1.5$\%. There is no difference in the Autoencoder model either with large or small number of classes.
We also examine dependence of ROC AUC, PR AUC on the lengths of visits sequence. They are in Figures \ref{Figure:roc_len} and \ref{Figure:pr_len} respectively.


In Table~\ref{bas} we present a comparison with the Isolation Forest Algorithm based on Word2Vec embeddings of tokens for treatments sequences.
Since the algorithm returns class labels, we compare recall and precision values.

\section{Conclusion}
\label{sec:conclusion}

We have investigated the unsupervised anomaly detection problem. The applied problem is from healthcare insurance, and the data is semi-structured sequences. 

We present unsupervised anomaly detection algorithms for semi-structured data never used before in the healthcare industry and compared them. 
Moreover, we propose an approach to natural normalization of errors based on the Empirical Distribution Function for better handling class imbalance within tokens. On top of these errors, we examine various aggregation strategies to provide a single anomaly score for a sequence. 

The overall quality of anomaly detection is similar for various LSTM and Autoencoder models and a various number of classes. Both models outperformed reasonable baselines, and thus provide a new baseline.
The usage of normalization further increases the quality of the LSTM model.

\section{Acknowledgments}
We thank Martin Spindler for providing the data and Ivan Fursov for providing code for data processing.

This work was supported by the federal program “Research and development in priority areas for the development of the scientific and technological complex of Russia for 2014–2020” via grant RFMEFI60619X0008.

\bibliographystyle{plain}
\bibliography{refs}

\end{document}